\documentclass{article}
\usepackage{ijcai22}

\usepackage{times}
\usepackage{soul}
\usepackage{url}
\usepackage[hidelinks]{hyperref}
\usepackage[utf8]{inputenc}
\usepackage[small]{caption}
\usepackage{graphicx}
\usepackage{amsmath}
\usepackage{amsthm}
\usepackage{booktabs}
\usepackage{algorithmic}
\usepackage[ruled]{algorithm2e}
\usepackage{multicol}
\usepackage{multirow}

\usepackage{bm}
\usepackage{graphicx}
\usepackage{subfigure}
\usepackage{color}

\usepackage{amsmath}
\usepackage{amsfonts}

\usepackage{float}

\theoremstyle{definition}
\newtheorem{definition}{Definition}

\title{Federated Learning  with GAN-based Data Synthesis for Non-IID Clients}

\author{
Zijian Li$^1$\and
Jiawei Shao$^2$\and
Yuyi Mao$^1$
\and
Jessie Hui Wang$^{3}$ \and
Jun Zhang$^{2}$\\
\affiliations
$^1$The Hong Kong Polytechnic University\\
$^2$The Hong Kong University of Science and Technology\\
$^3$Tsinghua University\\
\emails
zijian1997.li@connect.polyu.hk,
jiawei.shao@connect.ust.hk,
yuyi-eie.mao@polyu.edu.hk,
jessiewang@tsinghua.edu.cn, 
eejzhang@ust.hk
}

\begin{document}

\maketitle

\begin{abstract}
    Federated learning (FL) has recently emerged as a popular privacy-preserving collaborative learning paradigm. However, it suffers from the non-independent and identically distributed (non-IID) data among clients. In this paper, we propose a novel framework, named \emph{Synthetic Data Aided Federated Learning} (SDA-FL), to resolve this non-IID challenge by sharing synthetic data.
    Specifically, each client pretrains a local generative adversarial network (GAN) to generate differentially private synthetic data, which are uploaded to the parameter server (PS) to construct a global shared synthetic dataset.
    To generate confident pseudo labels for the synthetic dataset, we also propose an iterative pseudo labeling mechanism performed by the PS.
    A combination of the local private dataset and synthetic dataset with confident pseudo labels leads to nearly identical data distributions among clients, which improves the consistency among local models and benefits the global aggregation.
    Extensive experiments evidence that the proposed framework outperforms the baseline methods by a large margin in several benchmark datasets under both the supervised and semi-supervised settings.
\end{abstract}

\section{Introduction}


Federated learning (FL) is a privacy-aware paradigm that allows the clients to collaborate to learn a global model without sharing their local data \cite{non_iid_survey}.
Particularly, \cite{fedavg} introduced the Federated Averaging (FedAvg) algorithm where the clients train the local models based on the private local data and upload the model updates to the parameter server (PS) for aggregation.

Despite its success in the independent and identically distributed (IID) scenarios, FL still suffers from significant performance degradation when the data distribution among clients becomes skewed.
In particular, different clients learn from different data distributions in the non-IID scenarios, which leads to high inconsistency among the local models and thus degrades the effectiveness of global model aggregation \cite{FedNova}.

Many works have been proposed to alleviate the non-IID issue by regularizing the local models with the information of the global model and local models from other clients \cite{fedprox,scaffold}.
These methods, however, aim to reduce the local model bias and cannot achieve a significant improvement in scenarios with extreme non-IIDness \cite{fl_experiments}.
Recent studies have also attempted to tackle the non-IID problem with data augmentation techniques \cite{non_iid_survey}.
Specifically, \cite{fedmix,mix2fld} proposed to generate synthetic samples by mixing the real samples.
Nevertheless, without implementing a privacy-protection mechanism, these methods are susceptible to data leakage.

Observing the data heterogeneity problem and the privacy leakage of the existing data augmentation methods for FL, we propose a novel framework, named \emph{Synthetic Data Aided Federated Learning} (SDA-FL), which resolves the non-IID issue by sharing the differentially private synthetic data.
In this framework, each client pretrains a local differentially private generative adversarial network (GAN) \cite{gan} to generate synthetic data, thus avoiding sharing the raw data.
These synthetic data are then collected by the PS to construct a global synthetic dataset. 
To generate confident pseudo labels for the synthetic data, we propose an iterative pseudo label update mechanism, in which the PS utilizes the received local models to update the pseudo labels in each training round.
As the local models are progressively improved over the FL process, the confidence of pseudo labels is thus enhanced, which is beneficial for the server updates and local updates in future rounds and in turn results in a well-performed global model.
It is worth noting that the SDA-FL framework is compatible with many existing FL methods and can be applied in both supervised and semi-supervised settings without requiring labels of the real data, which will be validated in the experiments.
Ablation studies are also conducted to illustrate the impact of the privacy budget and the effectiveness of the key procedures in SDA-FL.
\section{Related Works}
\paragraph{Non-IID Challenges in Federated Learning}
The non-IID data distribution has been a fundamental obstacle for FL \cite{non_iid_survey}.
This is because the highly skewed data distribution significantly enlarges the local model divergence and thus deteriorates the performance of the aggregated model \cite{fedavgnoniid,FedRS}.
To mitigate the client drift caused by the non-IID data, many works proposed to modify the local objective function with the additional knowledge from the global model and local models of other clients \cite{fedprox,scaffold}.
Such methods, however, cannot achieve satisfactory performance in many non-IID scenarios \cite{fl_experiments}.
In addition to training the same model structure at clients, some studies proposed to combat the negative impact of non-IID data by adjusting the local model structures at individual clients \cite{fedpre,Think_locally_act_globally}.
Moreover, another thread of research addressed the data heterogeneity problem by optimizing the operations at the PS, including model aggregation \cite{FedNova}, client selection \cite{reinforcement_noniid,client_selection}, client clustering \cite{CFL,cluster_3}, and classifier calibration \cite{classifier_calibration}.

\paragraph{Data Augmentation and Privacy Preserving}

Recently, FL methods based on some form of data sharing have received increasing attention for their prominent performance \cite{non_iid_survey,fedavgnoniid}. 
A popular approach is to leverage the Mixup technique \cite{mixup} for data augmentation, so that the clients can share the blended local data and collaboratively construct a new global dataset to tackle the non-IID issue  \cite{mix2fld,xormmixup,fedmix}.
However, frequent data exchange may be vulnerable to privacy attacks.
Alternatively, GAN-based data augmentation \cite{fedavgnoniid,gan_non_iid,hybrid_fl} was shown to be effective in reducing the degree of local data imbalance in FL. 
The general idea is to train a good generative model at the server based on a few seed data samples uploaded by the clients.
Then this well-trained generator is downloaded by all clients for local model updating.
Nevertheless, since sending local data samples to the server violates the data privacy requirement, FedDPGAN \cite{GAN_FL_COVID19} suggested all the clients collaboratively train a global generative model based on the FL framework to supplement the scarce local data.
Unfortunately, the GAN training process also requires frequent exchange of generative models, leading to extremely high communication costs and risks of adversarial attacks \cite{Gan-leaks}.

\begin{figure*}
\setlength\abovecaptionskip{-0.2pt}
\setlength\belowcaptionskip{-5pt}
    \centering
    \includegraphics[width=0.7\textwidth]{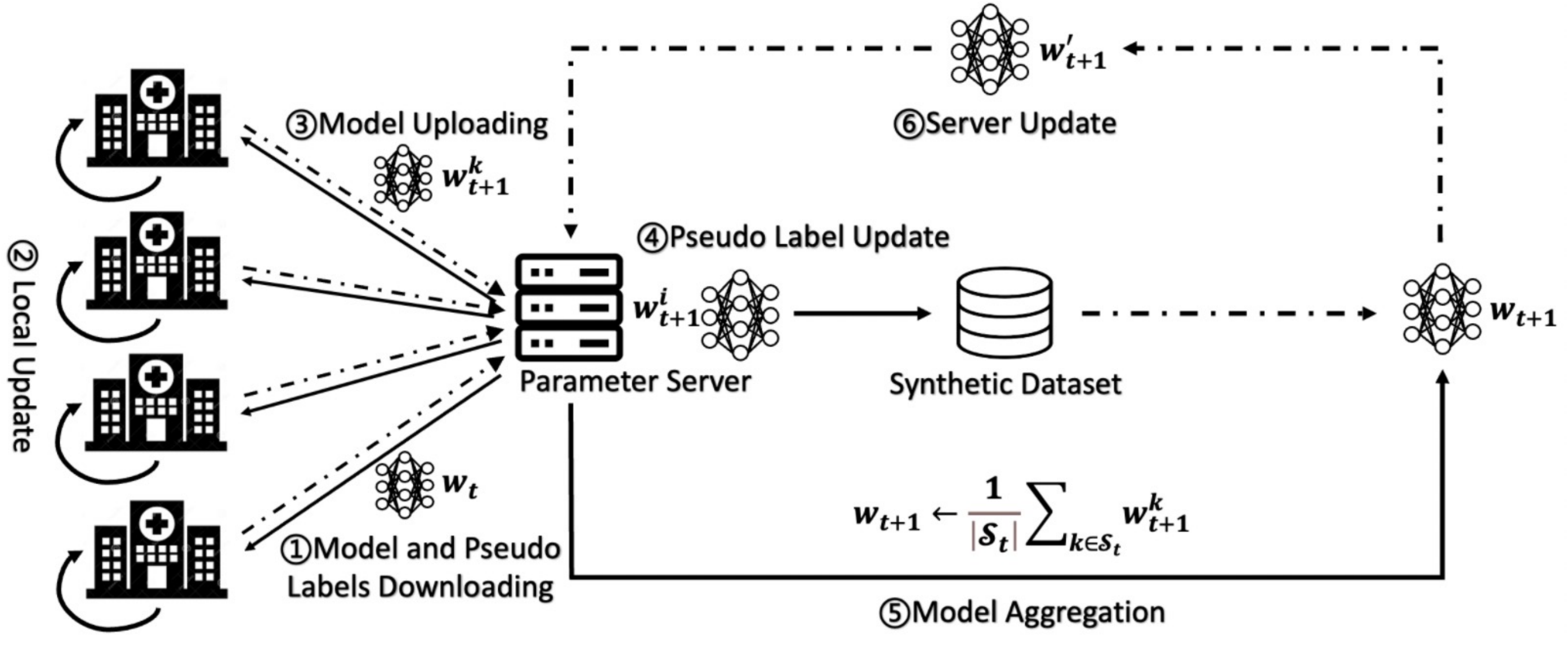}
    \caption{Overview of the proposed SDA-FL framework. Before the FL process starts, the synthetic data from all clients are sent to the PS to construct a global synthetic dataset. In each communication round, every client first downloads the global model and updates the pseudo labels of the synthetic data for local training.
The local models are then uploaded to the PS for pseudo label updating and model aggregation.
Lastly, the PS updates the global model ${\bm{w}_{t+1}}$ with the updated synthetic dataset.}
    \label{fig:System_Overview}
\end{figure*}

\section{Preliminary}

\paragraph{Federated Learning}

FedAvg \cite{fedavg} is a representative training algorithm for FL.
In each training round $t = 0,1,\dots,T-1$, every client in set ${ \mathcal{S}_t}$ downloads the global model $\bm{w}_t$ and updates the local model with the local dataset $\mathcal{D}_k=(\mathbf{X}_k, \mathbf{Y}_k)$ via stochastic gradient descent (SGD), i.e., $\bm{w}_{t+1}^k \leftarrow \bm{w}_t^k - \eta_{t} \nabla \ell(\bm{w}_t^k; \mathcal{D}_k)$, $k\in\mathcal{S}_{t}$, where $\mathcal{S}_{t}$ is a subset of clients activated in round $t$ and $\ell(\cdot)$ is the cross-entropy loss. The updated local models are then sent back to the PS for weighted aggregation $\bm{w}_{t+1} \leftarrow \frac{1}{|\mathcal{S}_t|} \sum_{k \in \mathcal{S}_t} \bm{w}_{t+1}^k$. These procedures repeat until all the $T$ training rounds are exhausted.

With the iterative local training and global aggregation procedures, the PS is expected to obtain a well-performed global model even without access to any local data. 
However, the highly skewed data distribution among clients easily leads to severe local model divergence and consequently degrades the global model performance \cite{fl_experiments}.
To solve this issue and avoid sharing the real data, we exploit the generative adversarial network (GAN) to generate high-quality synthetic data that can be shared among clients and used to update the local models and global model.

\paragraph{Differentially Private Generative Adversarial Network}

To avoid the gradient vanishing and mode collapse problems encountered by conventional GAN models \cite{wasserstein}, the Wasserstein GAN with gradient penalty (WGAN-GP) \cite{wgan-gp}, which penalizes the gradient norm of the critic to stabilize the training process of the generator $G$ and discriminator $D$, is adopted. 
With the real data distribution $p_r(x)$ and input noise distribution of the generator $p_z(z)$, the objective function of the WGAN-GP is expressed as follows:
\begin{equation}
\begin{split}
    \min_G {\max_{D \in \mathcal{D}} \mathbb{E}_{x \sim p_r(x)}\left[ D(x) \right] -\mathbb{E}_{z \sim p_z(z)} \left[ D \left( G(z) \right) \right]} \\
    {+ \gamma \left \| \nabla_{\hat{x}} {D(\hat{x})-1} \right \|_2^2}, 
\end{split}
\end{equation}
where $\mathcal{D}$ is the set of all $L$-Lipschitz functions, $\hat{x}$ is a mixture of the real sample $x$ and the fake sample $G(z)$, and $\gamma$ is a hyper-parameter.

To provide differential privacy protection for the synthetic data, we inject Gaussian noise into the GAN training process. The definition of differential privacy (DP) is given as follows:

\begin{definition}
(Differential privacy \cite{dp}): A random mechanism {$\mathcal{A}_{p}$} satisfies ($\epsilon$, $\delta$)-differential privacy if for any output's subset ($\mathcal{S}$) and any two adjacent datasets $\mathcal{M}$, $\mathcal{M}^{\prime}$, the following probability inequality holds:
\begin{equation}
    \mathbb{P}\left(\mathcal{A}_{p}(\mathcal{M}) \in \mathcal{S}\right) \leq e^{\epsilon} \cdot \mathbb{P}\left(\mathcal{A}_{p}\left(\mathcal{M}^{\prime}\right) \in \mathcal{S}\right)+\delta,
\end{equation}
\end{definition}
\noindent where $\delta > 0$ and $\epsilon$ is the privacy budget indicating the privacy level, i.e., a smaller value of $\epsilon$ implies stronger privacy protection.

To satisfy the $(\epsilon,\delta)$-DP, we follow \cite{dpgan} and add Gaussian noise to the updated gradients at each discriminator training iteration.
The relationship between the noise variance and differential privacy is shown as follows:
\begin{align}
\label{equ:sigma_epsilon}
    \sigma_{n}= \frac{2 q}{\epsilon} \sqrt{n_{d} \log \left(\frac{1}{\delta}\right)},
\end{align}
where $q$ and $n_d$ denote the sample probability for each instance and the total batch number of the local dataset, respectively.
Besides, according to the post-processing property of DP \cite{dp}, any mapping from the differentially private output also satisfies the same-level DP. In other words, the gradients of the generator, which are obtained via the backpropagation from the noisy discriminator output, also meet the $(\epsilon,\delta)$-DP.

Based on the training algorithm of WGAN-GP, it is desirable to train the generators with the label information, e.g., Auxiliary Classifier Generative Adversarial Network (ACGAN) \cite{ACGAN}, so that the synthetic data can be generated with labels. 
However, considering the label scarcity problem in the federated semi-supervised settings, this form of the conditional generative model is unable to be trained at clients.
Therefore, we resort to a more general paradigm that trains an unsupervised generator, and propose a pseudo labeling procedure within the FL process to generate high-confidence pseudo labels.

\paragraph{Pseudo Labeling}
To generate confident pseudo labels, following \cite{fixmatch}, only the class with an extremely high prediction probability is regarded as the pseudo label. 
Specifically, with a predefined threshold $\tau$, class $c$ is deemed as the label of sample $x$ if the output prediction probability $f_c(\bm{w};x)$ is the largest among all the classes and also larger than the threshold $\tau$. Hence, the pseudo labeling function can be expressed as follows:
\begin{equation}
\label{pseudo label}
    \hat{y}= \begin{cases}
        c &  \text{if } \max_{c} f_{c}(\bm{w};x) > \tau,\\
        \text{None} & \text{otherwise}.
    \end{cases}
\end{equation}
With such a pseudo labeling procedure, only the high-quality synthetic data can output a high prediction probability by model $\bm{w}$ and obtain their pseudo labels. As such, the unqualified synthetic samples are filtered, leaving only the qualified ones to update the local models.

In our proposed FL framework, the local models are used to predict the pseudo labels for the synthetic data generated by the corresponding local generators, and the pseudo labels are continuously updated with the improved local models during the FL process, as will be discussed in the next section.

\begin{table*}[t]
\setlength\abovecaptionskip{-0.1pt}
\setlength\belowcaptionskip{-5pt}
\centering
\scalebox{0.8}{
\begin{tabular}{ccccccc}
\hline
\textbf{Hospital}              & \multicolumn{1}{c}{\textbf{0}} & \multicolumn{1}{c}{\textbf{1}} & \multicolumn{1}{c}{\textbf{2}} & \multicolumn{1}{c}{\textbf{3}} & \multicolumn{1}{c}{\textbf{4}} & \multicolumn{1}{c}{\textbf{5}} \\ 
\cmidrule(lr){1-1} \cmidrule(lr){2-7}
\multirow{2}{*}{\textbf{Data}} & \multicolumn{1}{c}{Normal 2,000}            & \multicolumn{1}{c}{COVID-19 750}       & \multicolumn{1}{c}{Pneumonia 250}         & \multicolumn{1}{c}{Normal 2,000}        & \multicolumn{1}{c}{COVID-19 750 }         & \multicolumn{1}{c}{Pneumonia 250}           \\
& \multicolumn{1}{c}{COVID-19 750}            & \multicolumn{1}{c}{Pneumonia 250}         & \multicolumn{1}{c}{Normal 2,000}           & \multicolumn{1}{c}{COVID-19 750 }          & \multicolumn{1}{c}{Pneumonia 250}          & \multicolumn{1}{c}{Normal 2,000}           \\ \cline{1-7}
\end{tabular}}
\caption{The distribution of the COVID-19 dataset.}
\label{the COVID-19 experiment setting}
\end{table*}

\begin{table*}
\setlength\abovecaptionskip{-0.2pt}
\setlength\belowcaptionskip{-5pt}
\centering
\resizebox{\textwidth}{!}{
\begin{tabular}{ccccccccccccc}
\cmidrule(lr){1-13}
\textbf{\#class/client}
&\multicolumn{4}{c}{\textbf{\uppercase\expandafter{1}}}&\multicolumn{4}{c}{\textbf{\uppercase\expandafter{2}}}&\multicolumn{4}{c}{\textbf{\uppercase\expandafter{3}}}\\
\cmidrule(lr){1-1} \cmidrule(lr){2-5} \cmidrule(lr){6-9} \cmidrule(lr){10-13} 
\textbf{Algorithm} & \textbf{MNIST}   & \textbf{FashionMNIST} & \textbf{CIFAR-10} & \multicolumn{1}{c}{\textbf{SVHN}} & \textbf{MNIST}   & \textbf{FashionMNIST} & \textbf{CIFAR-10} & \textbf{SVHN}& \textbf{MNIST}   & \textbf{FashionMNIST} & \textbf{CIFAR-10} & \textbf{SVHN}
\\ \cmidrule(lr){1-13}
FedAvg  & 83.44 & 16.50  &18.36 &\multicolumn{1}{c}{14.05}  & 97.61 & 73.50 & 61.28 & 81.11  & 98.42 & 82.47 &79.33 & \multicolumn{1}{c}{84.18} \\ 
FedProx & 84.17 & 57.14  & 11.24  &  \multicolumn{1}{c}{17.53 }& 97.55 & 75.76 &63.16 &86.28  & 98.38 & 83.43   & 79.54  & \multicolumn{1}{c}{92.15 } \\
SCAFFOLD & 25.39&56.80  & 12.81 &  \multicolumn{1}{c}{11.64}& 94.17 & 70.82 & 60.78 & 73.34  &96.89  & 77.68  & 79.35  & \multicolumn{1}{c}{80.13 }\\
Naivemix & 84.35  & 66.62  & 14.39 &  \multicolumn{1}{c}{14.35}& 84.35&79.54 &64.39 &84.64 & 98.11  & 82.09  & 78.92  &  \multicolumn{1}{c}{92.30 }\\
FedMix   & 90.96  & 72.11  & 13.57 &  \multicolumn{1}{c}{16.78}& 90.96 & 82.41 &65.76 &86.61  & 98.46  & 84.65  & 79.49 &  \multicolumn{1}{c}{92.61 }\\
\textbf{SDA-FL}  &\multicolumn{1}{c}{\textbf{98.19}}   &\multicolumn{1}{c}{\textbf{85.70}} & \multicolumn{1}{c}{\textbf{37.70}} &  \multicolumn{1}{c}{\textbf{88.46}} & \multicolumn{1}{c}{\textbf{98.26}}  &\multicolumn{1}{c}{\textbf{86.87}} & \multicolumn{1}{c}{\textbf{67.89}} & \multicolumn{1}{c}{\textbf{90.70}}& \multicolumn{1}{c}{\textbf{98.50}  } &\multicolumn{1}{c}{\textbf{87.06} } & \multicolumn{1}{c}{\textbf{84.56} } & \multicolumn{1}{c}{\textbf{93.16} }
\\ \cmidrule(lr){1-13}
\end{tabular}}
\caption{Test accuracy (\%) of different methods on various datasets. 
}
\label{supervised experiment}
\end{table*}

\begin{table*}[!htp]
\setlength\abovecaptionskip{-0.2pt}
\setlength\belowcaptionskip{-5pt}
    \centering    
    
    \scalebox{0.8}{
    \begin{tabular}{ccccccccc}
    \cmidrule(lr){1-9}
     \textbf{Algorithm}    & 
     \textbf{FedAvg} &\textbf{FedProx} &\textbf{SCAFFOLD} & \textbf{Naivemix} & \textbf{FedMix} & \textbf{FedDPGAN} & \textbf{FedAvg (IID)} & \textbf{SDA-FL} \\
     \cmidrule(lr){1-1} \cmidrule(lr){2-9}
     \textbf{Accuracy} & 94.05 & 95.03 &94.30 & 94.14 & 94.28 & 94.57 & 95.19 & \textbf{96.25}\\
     \cmidrule(lr){1-9}
     
    \end{tabular}}
    \caption{Test accuracy (\%) on the COVID-19 dataset. FedAvg (IID) represents the scenario where the training samples are distributed to all the clients by average to achieve the IID distribution.}
    \label{COVID-19 experiment}
\end{table*}

\section{Synthetic Data Aided Federated Learning (SDA-FL)}

We now introduce the SDA-FL framework that adopts GAN-based data augmentation to alleviate the negative effect of the non-IID data.
The overview of the SDA-FL framework is shown in Figure \ref{fig:System_Overview}, and key algorithmic innovations built upon the classic FL framework are elaborated below.

\paragraph{Global Synthetic Dataset Construction}

At the start of the FL process, each client pretrains a local GAN model to generate synthetic samples based on its local data.
Then, the synthetic samples are sent to the PS to construct a global shared synthetic dataset.
To effectively leverage the synthetic dataset for FL, we perform pseudo labeling for these samples, which is critical for the effectiveness of the SDA-FL framework.

Unlike the prior work \cite{helpers} where each client leverages the local models from other clients to annotate the unlabeled data that encounters a bottleneck with highly skewed data distribution, we only utilize the local models to perform pseudo labeling for the corresponding unlabeled synthetic data.
This is because the local model and the corresponding synthetic data are trained with the same local data at each client, and only the high-quality synthetic samples can obtain a high prediction probability with this local model.
In addition, the confidence of the pseudo labels heavily relies on the local model quality, but the under-trained local models at the beginning of the FL process fail to accomplish this task.
Therefore, we update the pseudo labels for the global shared synthetic dataset with the stronger local models in each FL round.
Specifically, after receiving the local model ${\bm{w}_{t+1}^k}$ in round $t$, the PS assigns a pseudo label for each unlabeled synthetic instance $x$ according to \eqref{pseudo label}, i.e., its maximum class probability $f_c(\bm{w}_{t+1}^k;x)$ is higher than the predefined threshold $\tau$. In this way, we can gradually generate high-quality pseudo labels for the synthetic data samples.


\paragraph{Synthetic Data Aided Model Training}
Augmented by the samples $\hat{\mathbf{X}}$ and confident pseudo labels $\hat{\mathbf{Y}}^t$ from the shared synthetic dataset, the data available for local training at different clients are approximately homogeneously distributed. 
To make good use of the synthetic data, we leverage the Mixup method proposed in \cite{mixup},  which utilizes a linear interpolation between the real batch samples $(\mathbf{X}_{i,e}^t,\mathbf{Y}_{i,e}^t)$ and the synthetic batch samples $(\hat{\mathbf{X}}_e,\hat{\mathbf{Y}}_e^t)$, to augment the real data for client $i$ at round $t$:
\begin{equation}
\begin{aligned}
    \Bar{\mathbf{X}}_{i,e}^t & =\lambda_1 \hat{\mathbf{X}}_e+(1-\lambda_1) \mathbf{X}_{i,e}^t, \\ 
    \Bar{\mathbf{Y}}_{i,e}^t & =\lambda_1 \hat{\mathbf{Y}}_{e}^t+(1-\lambda_1) {\mathbf{Y}}_{i,e}^t,
\end{aligned}
\label{mixup}
\end{equation}
where $\lambda_1$ follows the Beta distribution $\mathrm{Beta}(\alpha,\alpha)$ for each batch with $\alpha \in [0,1]$.
By combining the cross-entropy loss $\ell(\cdot)$, the mixup loss for the local model update becomes:
\begin{equation}
    \ell_1 = \lambda_1 \ell \big(f(\Bar{\mathbf{X}}_{i,e}^t;\bm{w}_t),\hat{\mathbf{Y}}_e^t \big) + (1-\lambda_1) \ell \big(f(\Bar{\mathbf{X}}_{i,e}^t;\bm{w}_t),\mathbf{Y}_{i,e}^t \big).
\label{l1}
\end{equation}
In addition, since the loss in \eqref{l1} is fragile at the beginning of the FL process caused by the unconfident pseudo labels, another cross-entropy loss term is introduced for the real batch samples $(\mathbf{X}_{i,e}^t,\mathbf{Y}_{i,e}^t)$ to stabilize the training process:
\begin{equation}
    \ell_2 = \ell \big(f( \mathbf{X}_{i,e}^t;\bm{w}_t),\mathbf{Y}_{i,e}^t \big).
\end{equation}
Then, SGD is applied to update the local model as follows:
\begin{equation}
\label{equ:model_update}
    \bm{w}_{t+1}^k \leftarrow \bm{w}_t^k - \eta_{t} \nabla (\ell_1+\lambda_2 \ell_2),\;
\end{equation}
where $\lambda_2$ is a hyper-parameter to control the retention of the local data. 

In contrast to traditional FL where the PS does not have access to any data to update the global model, the PS in our framework keeps the entire global synthetic dataset $\hat{\mathcal{D}}_s$ and uses it to train the global model. 
Particularly, since there is no real data in the PS, two batches of synthetic samples are used to update the global model with \eqref{equ:model_update} at each iteration.

\paragraph{Interplay between Model Training and Synthetic Dataset Updating}
In each FL round, the aid of synthetic data improves the local models.
Since the updated local models are used for pseudo labeling and the global synthetic dataset construction at the PS, the confidence of the pseudo label is thus boosted. 
With the enhanced synthetic dataset, the PS can refine the global model and all the clients can improve their local models subsequently at the next round. 
Therefore, the interplay between model training and synthetic dataset updating at every training round is critical to achieving a well-performed global model.

\begin{figure*}
\setlength\abovecaptionskip{-0.2pt}
\setlength\belowcaptionskip{-5pt}
\centering  
\subfigure[MNIST]{
\label{Fig.semi_mnist}
\includegraphics[width=0.31\textwidth]{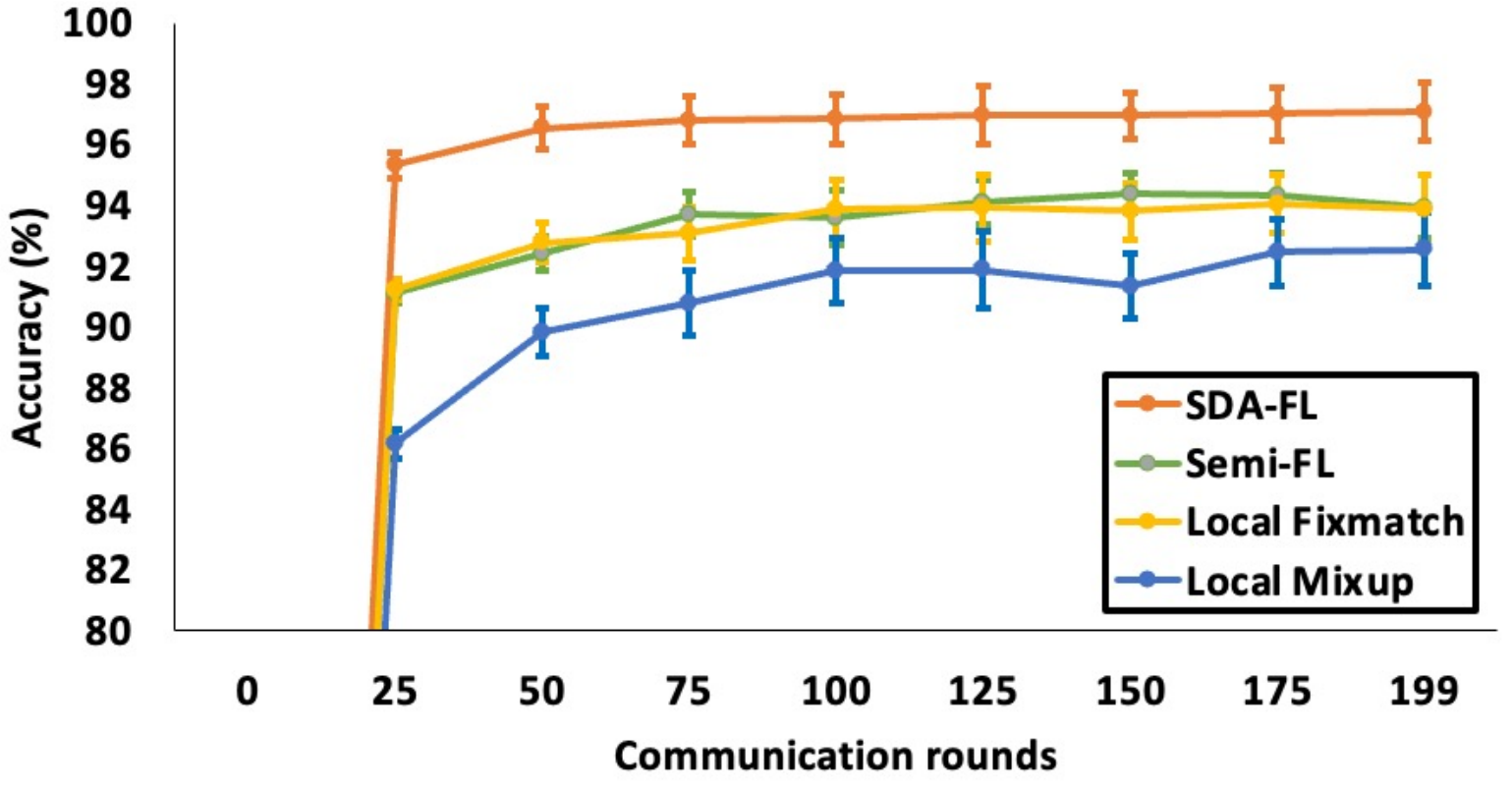}}
\subfigure[FashionMNIST]{
\label{Fig.semi_fashionmnist}
\includegraphics[width=0.31\textwidth]{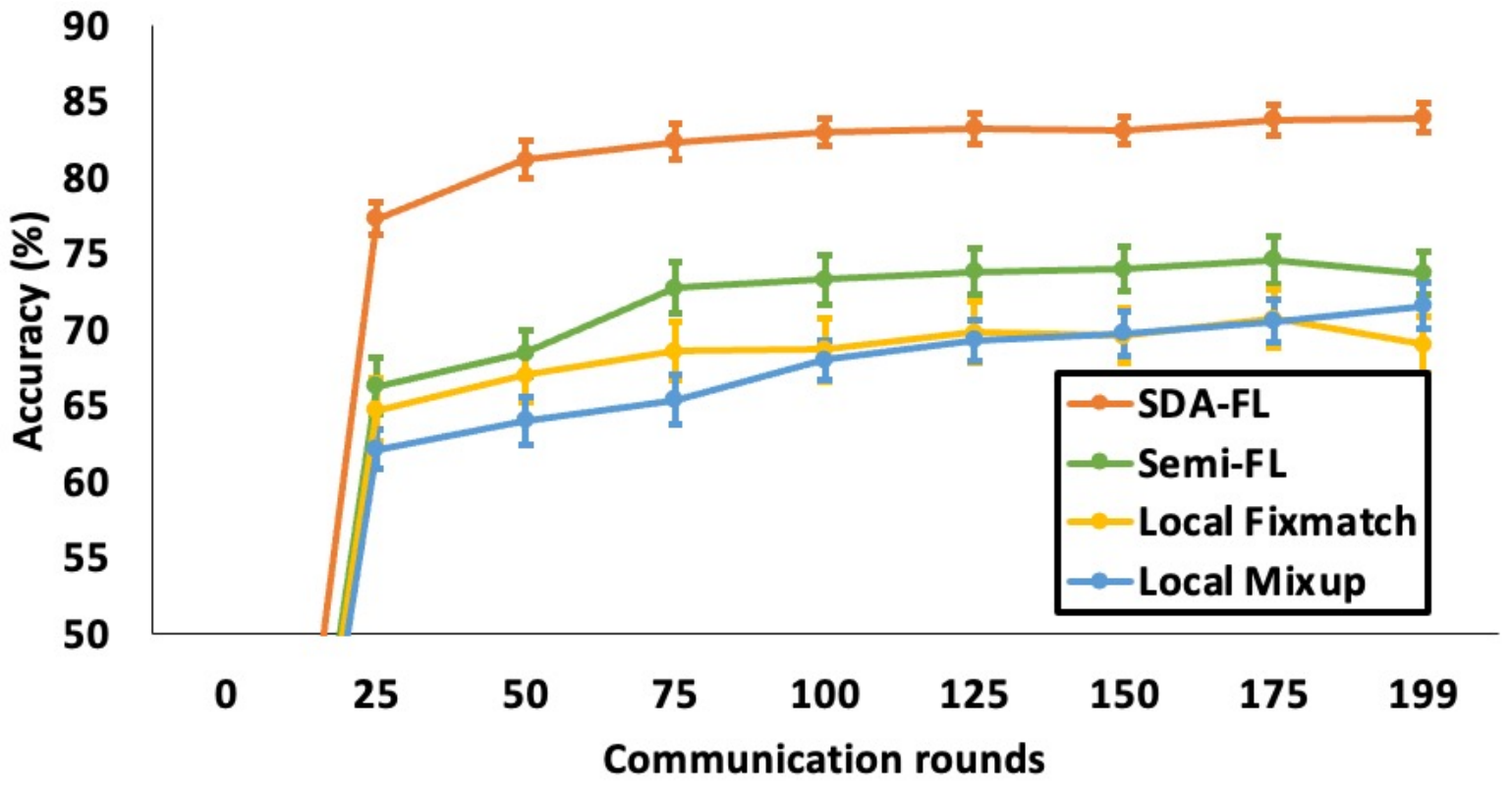}}
\subfigure[CIFAR-10]{
\label{Fig.semi_Cifar}
\includegraphics[width=0.31\textwidth]{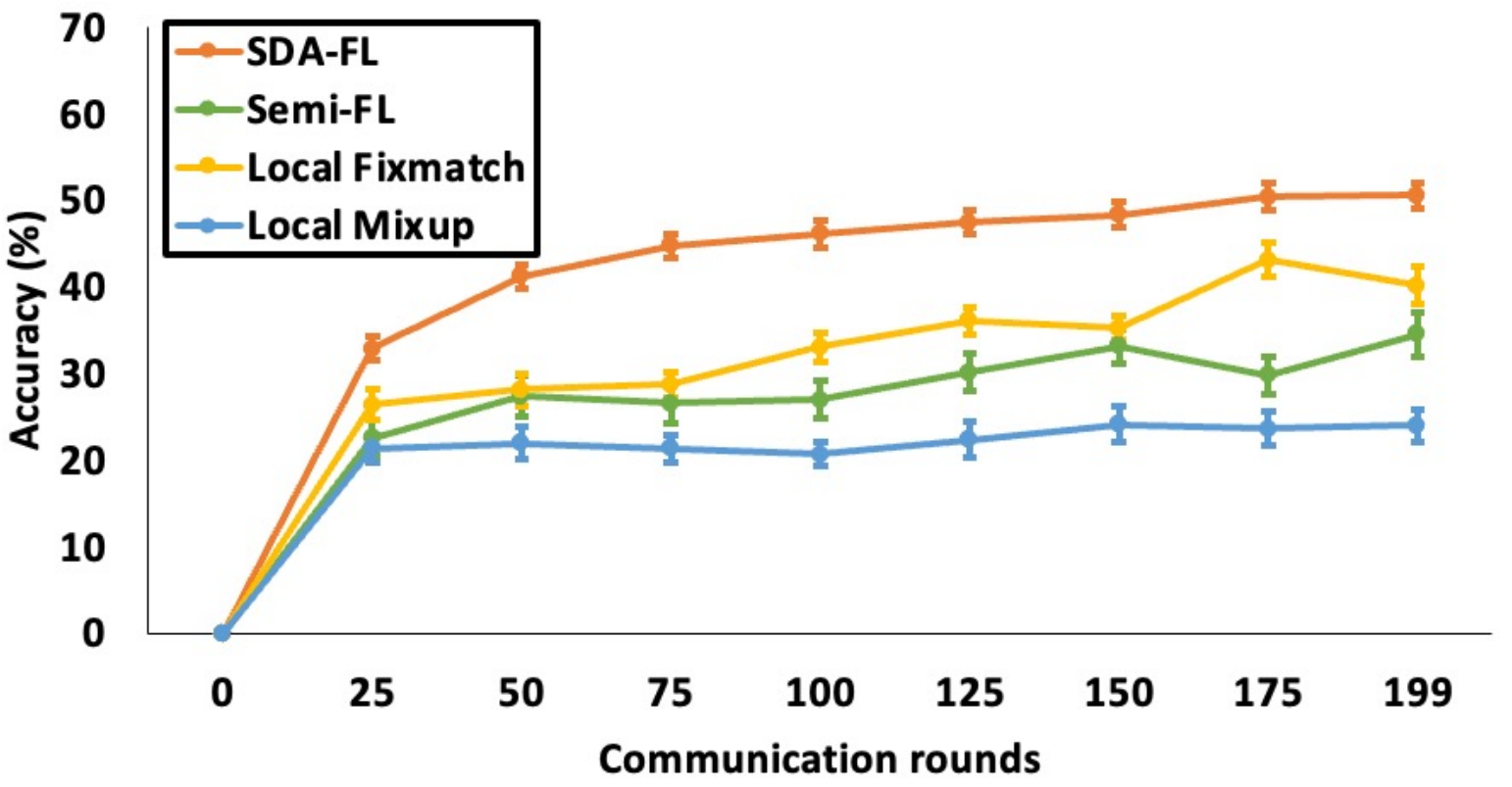}}
\caption{Test accuracy of different methods for federated semi-supervised learning on the MNIST, FashionMNIST, and CIFAR-10 datasets.}
\label{Fig.semi_experiment}
\end{figure*}

\paragraph{SDA-FL vs. Traditional FL}

From the above descriptions, the proposed SDA-FL framework introduces additional operations at both the clients and PS. These main innovations contribute to performance improvement with non-IID data.
In traditional FL algorithms \cite{fedprox,scaffold}, clients update their models based only on the local data, which easily leads to performance degradation when data are heterogeneous over clients.
In our framework, the local dataset is augmented by the GAN-based synthetic samples to alleviate the non-IID problem. 
Furthermore, the PS in traditional FL algorithms only performs simple model aggregation.
In contrast, the PS in the SDA-FL updates the global model with the high-confidence synthetic data, which further improves the global model performance.
Overall, with a shared synthetic dataset and an iterative pseudo labeling mechanism, SDA-FL overcomes the heterogeneous data distributions among clients and enhances the global model at the PS.
We envision that this framework can be extended to develop other data augmentation-based methods for both federated supervised learning and federated semi-supervised learning.

\section{Experiments}
In this section, we evaluate the proposed SDA-FL framework in the presence of non-IID data for both federated supervised learning and federated semi-supervised learning.
The experimental results on different benchmark datasets shall demonstrate the superiority of the proposed framework over the baseline methods.
Ablation studies are also conducted to illustrate the effectiveness of key procedures in SDA-FL.
\subsection{Experiment Setup}
\paragraph{Datasets}
We use four benchmark datasets, including MNIST \cite{mnist}, FashionMNIST \cite{fashionmnist}, CIFAR-10 \cite{cifar10}, and SVHN \cite{svhn}, to evaluate the proposed method.
In all the experiments, we equally divide the training samples and assign them to the clients.
Specifically, given the number of classes per client as $C$ and total $K$ clients, the whole training dataset is split into $K*C$ subsets, and each subset only has a single class.
Then all the subsets of data are randomly shuffled and distributed to the clients.
We assume two classes of FashionMNIST data at each client in the ablation studies.

To guarantee the effectiveness of SDA-FL, we set the hyper-parameters $\gamma$, $\lambda_2$, and threshold $\tau$ to be 10.0, 1.0, and 0.95, respectively.
We deploy ten clients in the experiments, and all of them are selected in each round.
To generate high-quality synthetic data, we pretrain the generator and discriminator with 36,000 iterations for CIFAR-10 and 18,000 for the other datasets in both the federated supervised experiments and federated semi-supervised experiments. Each client uploads 4,000 synthetic samples to the PS to construct the global synthetic dataset.
There are 200 rounds for all the methods. 
In each round of SDA-FL, the PS utilizes the synthetic dataset to update the global model with 10 iterations for CIFAR-10 and 50 iterations for the other datasets. 
Besides, we set the local step size $E=90$ in the federated supervised experiments and $E=40$ in the federated semi-supervised experiments, and select the SGD with learning rate $\alpha=0.03$ as the optimizer.
In each iteration, the clients update the local models with batch size $B=64$ for the federated supervised experiments and $B=80$, including 16 labeled samples and 64 unlabeled samples, for the federated semi-supervised experiments.

Moreover, to evaluate the proposed SDA-FL in practical applications, we also test all the methods on realistic COVID-19 dataset \cite{covid-19_dataset}.
Because of the scarcity of Pneumonia samples, we only assume six clients in this experiment, each of which has two classes of data as shown in Table \ref{the COVID-19 experiment setting}. 
We train the GAN models locally with 4,500 iterations, and update the local models with $30$ iterations and the global model with $10$ iterations.

\paragraph{Baselines}
For the federated supervised experiments, we compare the SDA-FL framework with FedAvg \cite{fedavg}, FedProx \cite{fedprox}, SCAFFOLD \cite{scaffold}, Naivemix, and FedMix \cite{fedmix} on the MNIST, FashionMNIST, CIFAR-10, SVHN, and COVID-19 datasets. 
For the COVID-19 dataset, we also adopt FedDPGAN \cite{GAN_FL_COVID19} for comparison, which trains a global GAN to solve the non-IID issue for medical applications.
We report the best results by tuning the hyperparameter $\mu$ of the regularization term for FedProx and the mixup ratio $\lambda$ for FedMix.
Besides, we extend our framework to the semi-supervised learning setting on MNIST, FashionMNIST, and CIFAR-10 datasets by performing pseudo labeling for the local unlabeled data. 
We compare the SDA-FL framework with SemiFL \cite{semifl}, Local Fixmatch \cite{fixmatch}, and Local Mixup \cite{mixup} to show its effectiveness.

\paragraph{Models}
We adopt a simple CNN model that consists of two convolutional layers and two fully-connected layers for the MNIST and FashionMNIST classification tasks.
Meanwhile, ResNet18 \cite{resnet} is used for classifying the CIFAR-10, SVHN, and COVID-19 datasets. To generate qualified synthetic samples, we use a generator with four deconvolution layers and a discriminator with four convolutional layers followed by a fully-connected layer. 

\begin{table*}[t]
\setlength\abovecaptionskip{-0.01pt}
\setlength\belowcaptionskip{-5pt}
\centering
\scalebox{0.8}{
\begin{tabular}{ccccccc}
\hline
  \textbf{Datasets}  & \multicolumn{2}{c}{\textbf{FashionMNIST}} & \multicolumn{2}{c}{\textbf{CIFAR-10 (2class/client)}}             & \multicolumn{2}{c}{\textbf{CIFAR-10 (3class/client)}} \\   \cmidrule(lr){1-1}\cmidrule(lr){2-3} \cmidrule(lr){4-5} \cmidrule(lr){6-7} 
  \textbf{Algorithms}  & \textbf{WGAN-GP  }          & \multicolumn{1}{c}{\textbf{AC-WGAN-GP}}          & \multicolumn{1}{c}{\textbf{WGAN-GP}} & \multicolumn{1}{c}{\textbf{AC-WGAN-GP}} & \multicolumn{1}{c}{\textbf{WGAN-GP}} & \multicolumn{1}{c}{\textbf{AC-WGAN-GP}} \\ \cmidrule(lr){1-1} \cmidrule(lr){2-3} \cmidrule(lr){4-5} \cmidrule(lr){6-7} 
\multicolumn{1}{c}{\textbf{FID}}           & \textbf{217.81}          & 220.39                                & \textbf{114.56}                    & 154.27                       & \textbf{129.25}                    & 162.16                       \\ 
\cline{1-1}
\multicolumn{1}{c}{\textbf{Accuracy (\%)}} & \textbf{83.76}           & 82.03                                 & \textbf{67.89}                     & 67.22                        & \textbf{84.56}                     & 83.53                        \\ \hline
\end{tabular}}
\caption{Test accuracy and FID comparison with AC-WGAN-GP on various datasets.}
\label{comparison_acwgan}
\end{table*}

\begin{figure*}
\setlength\abovecaptionskip{-0.2pt}
\setlength\belowcaptionskip{-5pt}
    \centering
    \subfigure[MNIST]{
    \label{Fig.privacy_mnist}
    \includegraphics[width=0.32\textwidth]{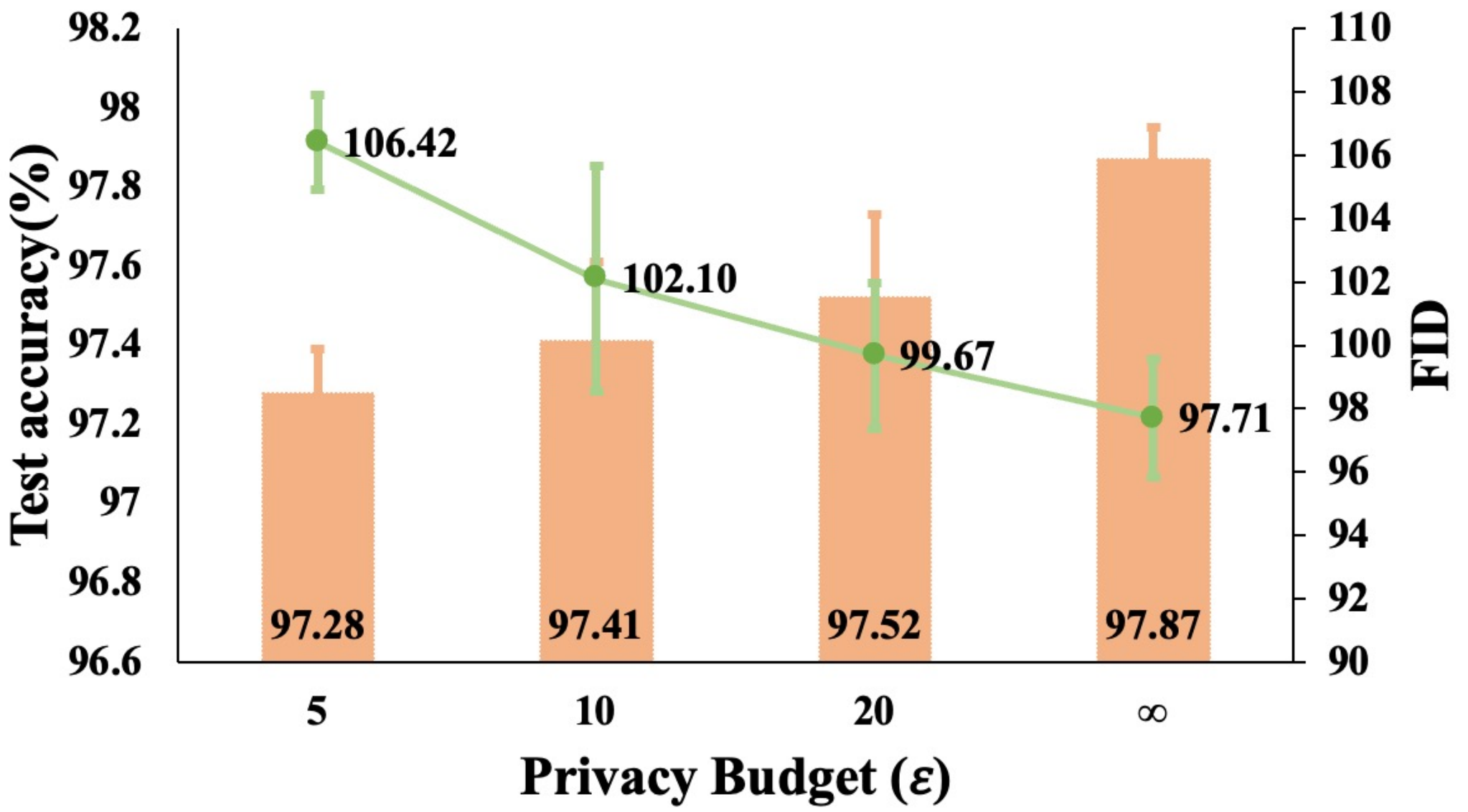}
    }
    \subfigure[FashionMNIST]{
    \label{Fig.privacy_fashionmnist}
    \includegraphics[width=0.32\textwidth]{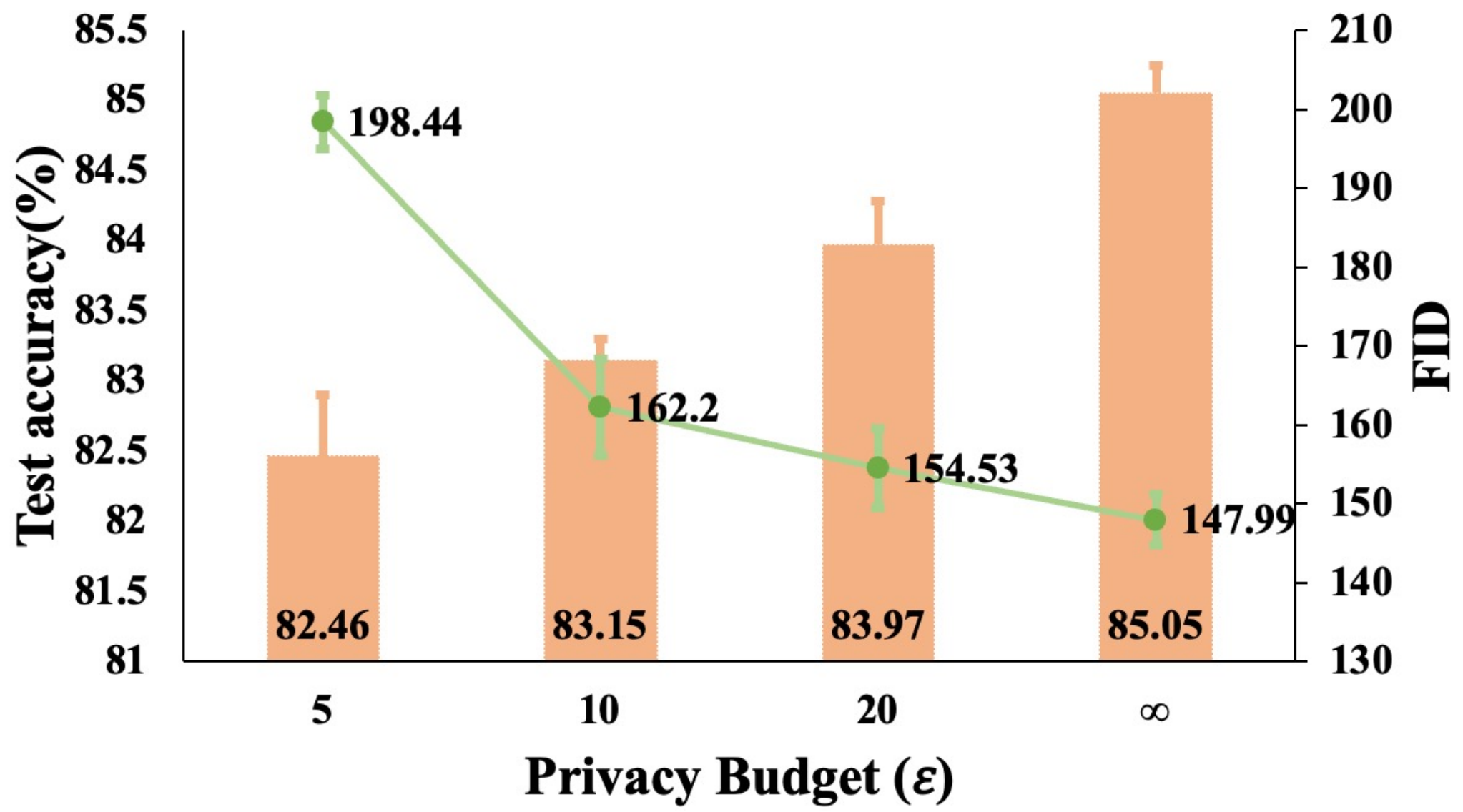}}
    \subfigure[CIFAR-10]{
    \label{Fig.privacy_cifar}
    \includegraphics[width=0.32\textwidth]{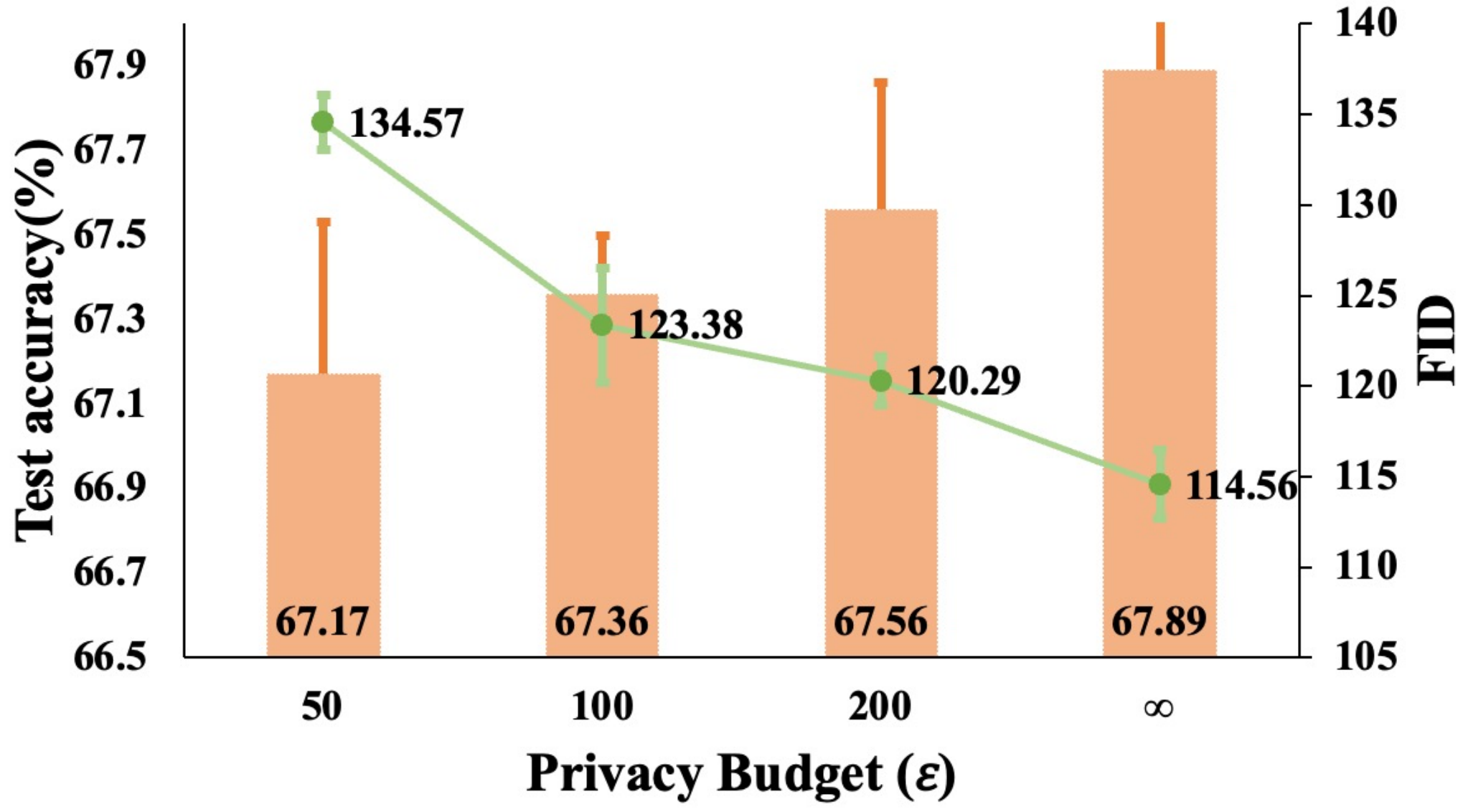}}
    \caption{Test accuracy and FID score with respect to the privacy budget. We run three trails and report the mean and the standard deviation of the test accuracy.
    The FID score of the real samples on MNIST, FashionMNIST, and CIFAR-10 are 10.54, 23.17, and 42.70, respectively, which are much larger than those of the synthetic data.}
    \label{Fig.privacy_level}
\end{figure*}


\subsection{Evaluation}
\paragraph{Performance in Federated Supervised Learning}
With varying numbers of classes per client, the experimental results in Table \ref{supervised experiment} show that our framework outperforms the baselines by a significant margin, which attributes to the GAN-based data augmentation that mitigates the detrimental effects of the data heterogeneity on FL.
In the CIFAR-10 experiments, our framework is superior to the Naivemix and FedMix algorithms at least by 5.0\% with three classes of data at each client, which verifies the competence of the GAN-based data compared with the mixing data.

In the COVID-19 experiments, besides the better performance over the above-mentioned baselines, SDA-FL also surpasses FedDPGAN by 1.68\% in accuracy. This demonstrates that the individually trained GANs generate synthetic data of higher quality than the global GAN trained based on the FL framework. Furthermore, in addition to resolving the non-IID issue, we find that SDA-FL even outperforms FedAvg (IID) by 1.14\% in accuracy, which shows its advantages for medical applications.

\paragraph{Performance in Federated Semi-Supervised Learning}
The results in Figure \ref{Fig.semi_experiment} show that the SDA-FL framework achieves faster convergence and better performance than other algorithms in the federated semi-supervised setting, indicating its robustness and generalizability. Particularly, compared with Semi-FL, our method improves the accuracy by almost 10\% on the FashionMNIST classification task.
In the CIFAR-10 dataset, the baseline methods are not able to train a usable global model (i.e., with a test accuracy below 40\%), while the proposed framework converges in this challenging scenario and improves the test accuracy significantly. 
This is because the proposed pseudo labeling mechanism can provide high-quality labels for the synthetic and unlabeled samples, which are beneficial to the FL process.

\begin{figure}
\setlength\abovecaptionskip{-0.2pt}
\setlength\belowcaptionskip{-5pt}
    \centering
    \subfigure[Server update]{
    \label{fig:server_update}
    \includegraphics[width=0.23\textwidth]{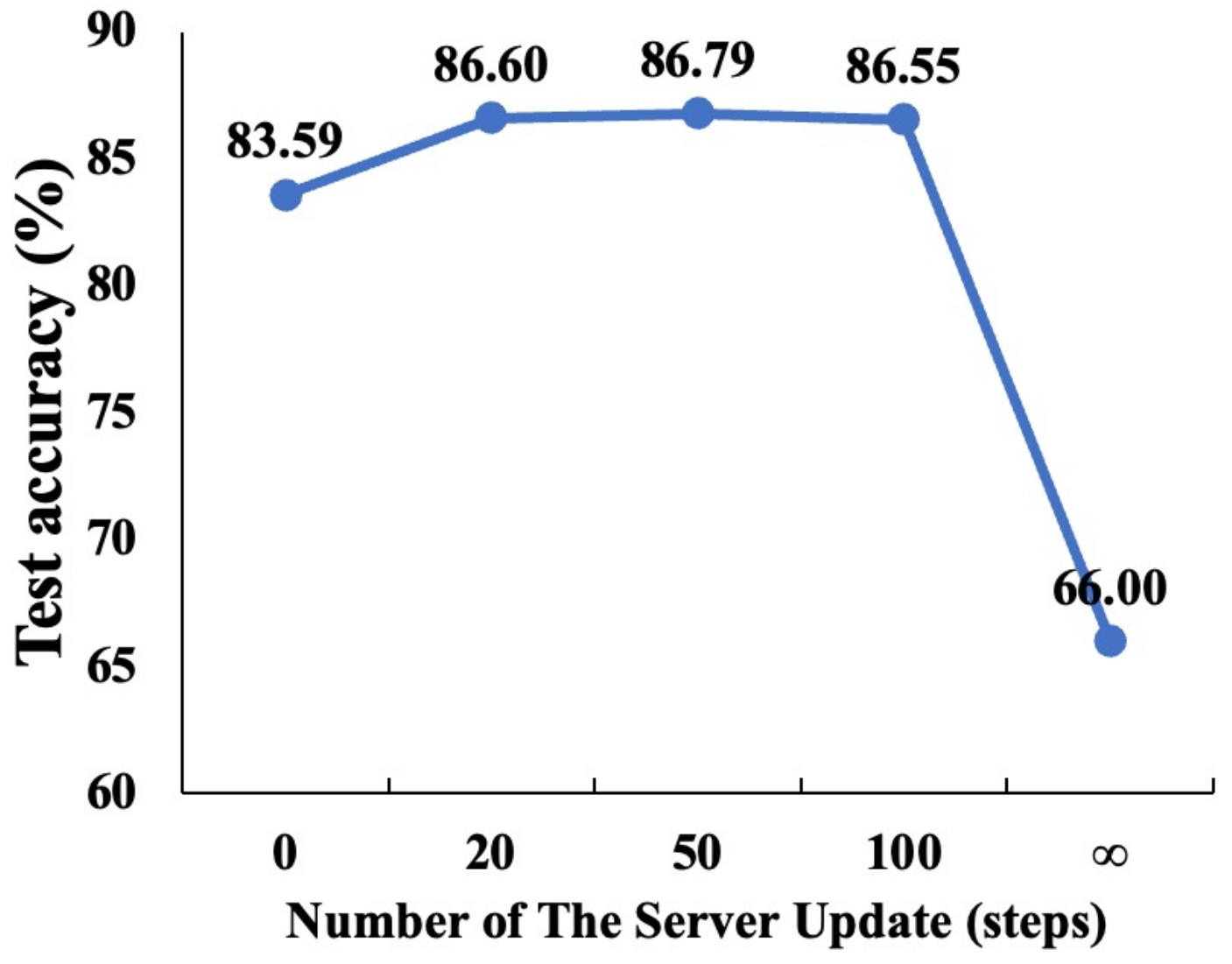}
    }
    \subfigure[Pseudo label update]{
    \label{fig:Pseudo_labeling}
    \includegraphics[width=0.23\textwidth]{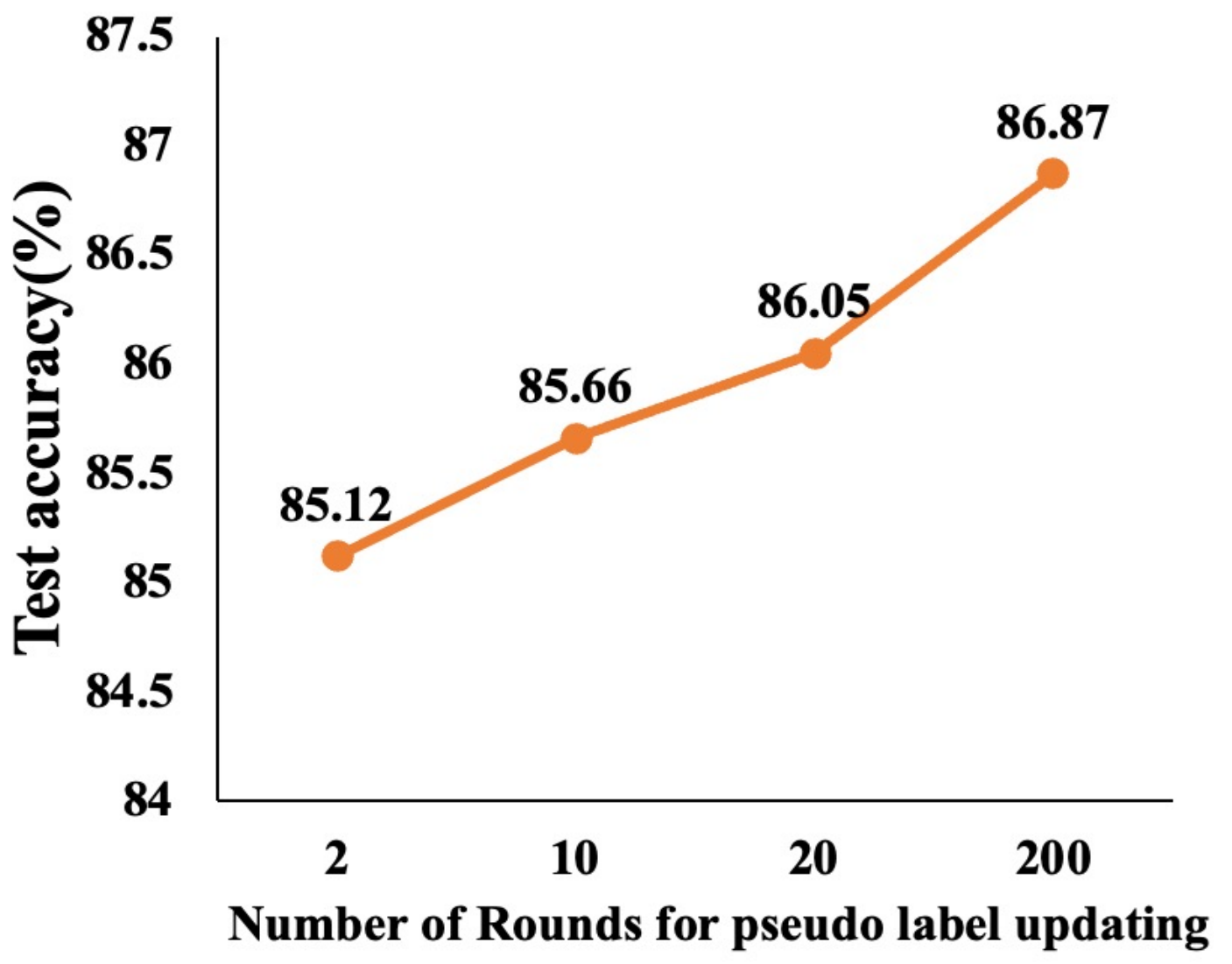}}
    \caption{Test accuracy on FashionMNIST with varying steps for server updating and rounds for pseudo label updating. The $\infty$ steps in (a) mean that the model is only trained with the synthetic data, and the 10 rounds in (b) represent that the PS only updates the pseudo labels in the first 10 rounds.}
    \label{Fig.server_update_pseudo_label}
\end{figure}



\paragraph{Tradeoff between the Privacy Budgets and Model Performance}
To investigate the impact of the privacy budgets, we evaluate the model performance of the SDA-FL framework under different values of $\epsilon$.
The Fréchet inception distance (FID) is used to measure the quality of the generated samples, where a smaller FID score indicates better image quality.
As illustrated in Figure \ref{Fig.privacy_level}, a strict privacy budget $\epsilon=5$ increases the FID score compared with that in the protection-free scenario, which implies quality degradation of the generated samples. 
This negative impact on the synthetic samples also leads to around 0.61\% and 2.59\% accuracy drop on the MNIST and FashionMNIST datasets, respectively.
The CIFAR-10 classification task follows a similar trajectory.
Please note that although the proposed SDA-FL framework is trained under strict privacy requirements, compared with the results in Table \ref{supervised experiment}, it still maintains supreme performance.


\paragraph{Effectiveness of Server Update and Pseudo Label Update} In comparison to the traditional FL, our framework updates the global model with the synthetic data at the PS, which has the potential to further improve the performance. 
The results in Figure \ref{fig:server_update} show that the model performance on FashionMNIST reduces by nearly 3\% without any server updates. 
Nevertheless, updating the global model too much by the PS may degrade the performance because of the excessive involvement of synthetic data.
Empirical results show that the model trained solely with the synthetic data (i.e., the server updates the global model for $\infty$ steps) can only obtain an accuracy of 66.0\%, which highlights the necessity of judicious utilization of the synthetic and local data for model training.

Besides, keeping updating pseudo labels in each round adopted by the SDA-FL framework for the synthetic data improves the model performance.
As illustrated in Figure \ref{fig:Pseudo_labeling}, the accuracy increases with the number of rounds for pseudo label updating, which demonstrates that the SDA-FL framework can improve the confidence level of the pseudo labels over the training process.
Note that since our framework only transmits the pseudo labels instead of the samples, the extra communication overhead is negligible.

\paragraph{Performance Comparison with Auxiliary Classifier WGAN-GP (AC-WGAN-GP)}
We can also include the label information in the GAN training to generate synthetic data with labels in the federated supervised experiments. 
As such, we compare the performance of SDA-FL with WGAN-GP and AC-WGAN-GP \cite{ACGAN} on the FashionMNIST and CIFAR-10 datasets. As shown in Table \ref{comparison_acwgan}, with the same number of training iterations for the generators, WGAN-GP achieves higher synthetic data quality as implied by the higher FID scores. 
Although AC-WGAN-GP can generate labeled synthetic data, WGAN-GP still performs better in accuracy performance with the higher-quality synthetic data. This is because our proposed pseudo labeling mechanism provides confident pseudo labels for the synthetic data.

\section{Conclusions and Discussions}

We proposed a new data augmentation method to resolve the heterogeneous data distribution problem in federated learning by sharing the differentially private GAN-based synthetic data. To effectively utilize the synthetic data, a novel framework, named Synthetic Data Aided Federated Learning (SDA-FL), was developed, which generates and updates confident pseudo labels for the synthetic data samples.
Experiment results showed that SDA-FL outperforms many existing baselines by remarkable margins in both supervised learning and semi-supervised learning under strict differential privacy protection.
In this study, we limit our attention to the WGAN-GP and AC-WGAN-GP in SDA-FL. 
Despite their performance improvements compared with the baselines, it is necessary to investigate other GAN structures.
In addition, WGAN-GP requires significant computational resources at clients. 
Therefore, to improve the applicability of SDA-FL, it is important to develop a computation-efficient GAN-based structure for clients in future research.




{\small
\bibliographystyle{named}
\bibliography{ijcai22}
}
\end{document}